\documentclass[10pt,twocolumn,letterpaper]{article}

\usepackage[pagenumbers]{cvpr} 

\usepackage{makecell}

\usepackage{cuted} 
\makeatletter
\newcommand{\cvprTeaser}[1]{\gdef\cvpr@teaser{#1}}
\def\cvpr@teaser{}
\providecommand{\linenumbers}{}
\providecommand{\nolinenumbers}{}
\apptocmd{\maketitle}{%
  \ifx\cvpr@teaser\@empty\else
    \nolinenumbers
    \begin{strip}
      \centering
      \vspace{-4mm}
      \cvpr@teaser
      \vspace{-4mm}
    \end{strip}
    \linenumbers
  \fi
}{}{}
\makeatother

\definecolor{cvprblue}{rgb}{0.21,0.49,0.74}
\usepackage[pagebackref,breaklinks,colorlinks,allcolors=cvprblue]{hyperref}

\def\paperID{    } 
\def\confName{CVPR}
\def\confYear{2026}

\title{Physics-Informed Video Flare Synthesis and Removal Leveraging Motion Independence between Flare and Scene}

\author{Junqiao Wang \quad Yuanfei Huang \quad Hua Huang\\
School of Artificial Intelligence, Beijing Normal University\\
{\tt\small 202321081032@mail.bnu.edu.cn \quad \{yfhuang, huahuang\}@bnu.edu.cn}
}

\usepackage[inkscapeexe=D:/program_file_self/Inkspace/bin/inkscape.exe]{svg}        

\cvprTeaser{%
  \includesvg[width=1.0\textwidth]{fig/motivation/motivation}
  \captionof{figure}{By modeling the motion independence between the flare and the scene, MIVF effectively removes video flare while maintaining the scene’s spatio-temporal consistency.}\label{fig:motivation}
}

\begin{document}
\maketitle
\begin{abstract}
Lens flare is a degradation phenomenon caused by strong light sources. Existing researches on flare removal have mainly focused on images, while the spatiotemporal characteristics of video flare remain largely unexplored. Video flare synthesis and removal pose significantly greater challenges than in image, owing to the complex and mutually independent motion of flare, light sources, and scene content. This motion independence further affects restoration performance, often resulting in flicker and artifacts. To address this issue, we propose a physics-informed dynamic flare synthesis pipeline, which simulates light source motion using optical flow and models the temporal behaviors of both scattering and reflective flares. Meanwhile, we design a video flare removal network that employs an attention module to spatially suppress flare regions and incorporates a Mamba-based temporal modeling component to capture long range spatio-temporal dependencies. This motion-independent spatiotemporal representation effectively eliminates the need for multi-frame alignment, alleviating temporal aliasing between flares and scene content and thereby improving video flare removal performance. Building upon this, we construct the first video flare dataset to comprehensively evaluate our method, which includes a large set of synthetic paired videos and additional real-world videos collected from the Internet to assess generalization capability. Extensive experiments demonstrate that our method consistently outperforms existing video-based restoration and image-based flare removal methods on both real and synthetic videos, effectively removing dynamic flares while preserving light source integrity and maintaining spatiotemporal consistency of scene.
Code is available at https://github.com/BNU-ERC-ITEA/MIVF.
\end{abstract}    
\section{Introduction}\label{sec:intro}

In digital imaging and computer vision, lens flare is a common degradation 
phenomenon caused by strong light sources. It typically manifests in images as 
halos, streaks, or saturated blobs, which significantly reduce image quality 
and undermine the reliability of high level vision tasks\cite{2024Flare7K}. 
For instance, in optical flow estimation, flare regions may be mistakenly 
recognized as foreground motion. In object detection\cite{huang2025deflaremamba}, 
intense flare spots can trigger false positives and erroneous tracking.

Hardware solutions\cite{Eric01} such as anti-reflective coatings or lens hoods can mitigate flare 
formation but cannot completely eliminate it. As a result, more practitioners are adopting algorithmic 
methods for flare removal, which serve as a less expensive alternative.
Traditional image processing methods\cite{seibert1985tradition01, asha2019auto, vitoria2019auto} can handle certain types of lens flare,  but their capability remains limited when dealing with diverse and complex flare phenomena. In recent years, deep learning has driven significant progress in image flare removal, with research focusing on dataset construction\cite{wu2021train, dai2022flare7k} and network design\cite{Song_2023_CVPR, ffswin_2023_cvpr}. 

However, most existing studies focus on images, and video flare removal has not yet been systematically explored. 
Compared to image flare removal tasks, modeling and removing flares in videos pose greater challenges, 
requiring a deeper understanding of the underlying physical properties. Specifically, as illustrated in Figure \ref{fig:motivation}, we focus on the motion independence between flare and scene.
In general scenes, scattering flares move consistently with the light source, 
they can be approximated as linearly following the motion of the light source.
In contrast, reflective flares exhibit motion independent of the scene, 
making their modeling and synthesis in real videos considerably more difficult.
Moreover, due to the motion independence between flares and scene content, 
most video restoration methods tend to entangle scene and flare features, 
thereby degrading restoration performance.

To address this challenge, we conduct the first systematic study on video flare modeling and removal.
We analyze the motion patterns of light sources and flares. The spatial distribution and morphology of lens 
flares are inherently correlated with the position and motion of light sources. 
Scattering flares typically move synchronously with the light source and are distributed around it, 
whereas reflective flares is independent of the scene but adheres to physical constraints tied to light source positions.

Based on this, we design a physics-informed dynamic flare synthesis pipeline.
In this new pipeline, simulated light sources are added into the scene, and their positions are updated according to the estimated optical flow.
For reflective flares, whose motion is independent from the scene, we model their motion behavior through physical constraints based on light source positions.

Moreover, inspired by recent progress in Mamba architectures, we further propose a Motion-Independent Spatio-Temporal Representation Network for Video Flare Removal (MIVF).
MIVF employs an attention mechanism to capture spatial flare characteristics and leverages Mamba's long range dependency mg to leodelinarn temporal dynamics.
Experimental results on the synthesized flare dataset show that MIVF significantly outperforms existing video restoration models in flare removal.
Moreover, evaluations on real-world video samples demonstrate that the proposed network, trained on our constructed dataset, effectively eliminates both scattering and reflective flares while maintaining high spatio-temporal consistency of the video scenes.

Our main contributions are summarized as follows.
\begin{itemize}
    \item We systematically analyze the formation mechanisms of scattering and reflective flares in videos, revealing the dynamic trait of video flares.

    \item We propose a physics-informed dynamic flare synthesis pipeline and construct a video flare dataset containing large scale synthesized videos, providing a strong foundation for video flare removal research.

    \item We design MIVF, a video flare removal network integrating attention-based spatial modeling with Mamba-driven temporal modeling, demonstrating superior performance and generalization to complex flare scenarios.
\end{itemize}

\section{Related Work}\label{related_work}
In this section, we review the related work in three perspectives: flare degradation modeling, 
flare removal methods, and video restoration.

\subsection{\bf Flare Degradation Modeling}
Among various low level vision tasks, flare removal has recently become a research hotspot 
due to its complex degradation patterns and the lack of sufficient paired training data. 
To construct large scale paired datasets of clean and corrupted images, researchers have 
analyzed the physical formation and visual characteristics of lens flare. Wu \etal~\cite{wu2021train} 
explicitly modeled the optical cause of flare using wave optics and generated semi-synthetic corrupted 
image pairs. Dai \etal~\cite{dai2022flare7k} designed a variety of flare components based on nighttime flare 
characteristics for synthetic data generation, and in a subsequent work introduced captured flare patterns 
in Flare7K++~\cite{2024Flare7K}, providing one of the most advanced flare datasets to date. 
Zhou \etal~\cite{zhou2025image} proposed a synthesis pipeline that aligns more closely with the ISP’s 
physical mechanism, while Qu \etal~\cite{qu2025flarex} employed a physics-based rendering approach to produce 
flare image pairs under both 2D and 3D perspectives.
However, these methods focus on static flare modeling, without considering the dynamic 
motion and variation of flares in videos.

\subsection{\bf Flare Removal Methods}
Early flare removal approaches relied on traditional image processing techniques, 
such as deconvolution~\cite{seibert1985tradition01} and light spot detection~\cite{asha2019auto}, 
which are effective only for limited flare types. With the development of deep learning and computer vision, 
learning-based methods have enabled more efficient and adaptive flare removal. Wu \etal~\cite{wu2021train} and 
Dai \etal~\cite{dai2022flare7k} first established data-driven training frameworks for flare removal. 
Subsequent researches further improved performance by incorporating flare 
priors~\cite{dai2023nighttime, deng2024towards, he2025disentangle}, enhancing generalization across different scenarios.
In addition, many studies have explored the use of adversarial training~\cite{zhou2025image}, 
light source modeling strategies~\cite{2024Flare7K}, and restoration network 
redesigns~\cite{Song_2023_CVPR, ffswin_2023_cvpr} to improve image flare removal.
Meanwhile, the rapid development of image restoration networks has introduced more powerful 
architectures. Recent Mamba-based models have shown impressive performance in image restoration tasks, 
and flare removal methods built upon Mamba~\cite{huang2025deflaremamba} currently achieve the 
state-of-the-art performance in this domain.

\subsection{Video Restoration}
Compared with image methods, video restoration theoretically benefits from temporal redundancy across frames, allowing improved degradation suppression and finer detail reconstruction.
However, this assumption does not hold for video flares, where the flare motion is complex and physically independent of the scene motion. 

Existing video restoration models developed for other tasks have demonstrated remarkable capability and thus provide an important reference for this work.
Most existing methods perform temporal modeling through frame alignment.
Early approaches relied on explicit image registration to establish temporal correspondence between neighboring frames~\cite{kappeler2016video, caballero2017real, tassano2020fastdvdnet}.
Subsequent methods introduced optical flow and deformable convolutions to achieve more accurate temporal alignment~\cite{chan2021basicvsr, wang2019edvr, chan2021basicvsr, isobe2020video}.

Presently, no dedicated approach or benchmark dataset has been proposed for video flare removal. Due to the motion independence between flare and scene, alignment-based strategies such as optical flow or deformable convolutions struggle to disentangle flare and scene dynamics, while long range recurrent or Transformer-based models incur heavy computational costs when processing extended sequences.
Therefore, video flare removal demands a new spatio-temporal modeling paradigm that can effectively handle the unique motion independence and dynamic behavior of video flares.
\section{Preliminaries}\label{sec:preliminaries}

\begin{figure}[t]
  \centering
  \includesvg[width = 1.0\linewidth, keepaspectratio]{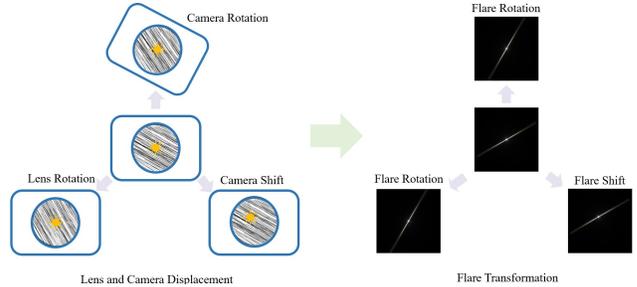} 
  \caption{Differences of Scattering Flares under Various Lenses and Camera Positions.
For each component, we simulate flare patterns under circular and hexagonal aperture shapes.
Within the same configuration, the effect of lens or camera rotation is modeled by controlling the orientation of the flare streaks.}\label{fig:pr_sc}
\end{figure}

Lens flares in images can be categorized into scattering flares and reflective flares.
Due to their distinct physical formation mechanisms, these two types exhibit significantly different visual characteristics.
In video sequences, their motion patterns and temporal variations also differ according to their respective physical properties.
In the following, we provide a detailed analysis of the spatial characteristics of these two flare types and their typical motion behaviors in dynamic scenes.

\subsection{Scattering Flare}
Scattering flare typically appears as streak or linear patterns, often caused by manufacturing imperfections or surface contamination on the lens, such as scratches and dust particles.
When light passes through these fine linear obstructions within the optical system, diffraction occurs, forming streaks perpendicular to the direction of the scratches.

As shown in Figure~\ref{fig:pr_sc}, scattered flare exhibits distinct characteristics 
in dynamic scenes. When the optical structure remains unchanged, \ie, 
neither the lens nor the barrel rotates, and the camera itself does not rotate the 
orientation of linear occluders within the optical system stays fixed. 
As the light source translates within the scene (ignoring minor caustic and distortion 
effects), the overall shape and direction of the scattered flare remain largely invariant.

\subsection{Reflective Flare}
Reflective flares typically appear as spots, circular, or polygonal halos in images. 
In an optical system, multiple lens elements are usually assembled to form a compound lens. 
When light enters the system and propagates through these lenses, a portion of the energy cannot be fully suppressed by the anti-reflective coatings. 
Consequently, partial reflections occur at each air-glass interface~\cite{koreban2009geometry}.

\begin{figure}[t]
    \centering
    \includesvg[width = 1.0\linewidth, keepaspectratio]{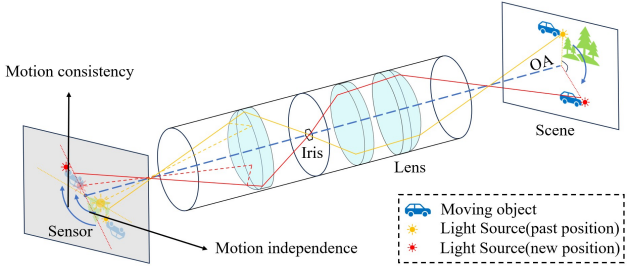} 
    \caption{The yellow and red lines represent light paths from the light source at different positions, while the long dashed lines denote the paths of the reflected flare light.
As the light source moves across the scene, the reflective flare does not move along with the scene. Instead, it remains physically constrained to stay collinear with the light source and the image center, as indicated by the short dashed lines.}\label{fig:rf}
\end{figure}

At these interfaces, the incident light is reflected outward from one internal surface and may subsequently reflect again from another, returning toward the image sensor. 
As illustrated in Figure~\ref{fig:rf}, such secondary reflections deviate from the path of the refracted light, forming distinct aperture artifacts on the image plane.

Even when considering only second-order reflections, if the optical system contains \( n \) lens elements, there exist \( 2n \) air-glass interfaces, and the number of possible second-order reflection paths can reach \( 2n^{2} - n \). 
These multiple reflection paths produce a series of circular or polygonal halos that are especially prominent in camera lenses.

When the camera lens features a large aperture and multiple optical groups, internal reflections occur more frequently. 
Each reflective flare is distributed along the line connecting the light source center and the optical axis (i.e., the image center). 
The position and morphology of these reflective artifacts vary with the incident angle of light \( \theta \). 
Therefore, the motion of the light source becomes the dominant factor influencing both the spatial distribution and geometric variation of reflective flares.

\subsection{Light Sources and Lens Flares}
The spatial morphology of lens flares are inherently correlated with the 
position and motion of light sources. As shown in Figure~\ref{fig:rf}
Scattering flares typically move synchronously with the light source and are 
distributed around it, whereas reflective flares is independent of the scene but adheres 
to physical constraints tied to light source positions.

In addition, through observation of multiple flare video sequences, we find that light source motion is generally consistent with that of foreground objects. 
This phenomenon can be categorized into two primary cases.
(1) when the camera remains stationary, the light source moves together with the object to which it is attached (e.g., vehicle headlights moving with the car). 
(2) when the light source and the scene remain static while the camera moves, the scene and foreground objects exhibit inverse motion relative to the camera, causing the apparent motion of the light source to follow the same inverse direction (e.g., the sun appearing to move oppositely as the camera pans). 
Therefore, the motion direction of the light source is consistent with that of the object or background to which it belongs.
\section{Method}

\subsection{Dynamic Flare Synthesis Pipeline}

\noindent\textbf{Scattering Flare.}
According to the theoretical analysis of flare dynamics described above, scattering flares generally remain stable in most scenarios. 
Therefore, we follow the simulation modeling strategy proposed by Wu \etal~\cite{wu2021train} to synthesize scattering flares. 

\begin{figure*}[t]
  \centering
  \begin{subfigure}[t]{0.50\textwidth}
    \centering
    \includesvg[height=4.50cm,keepaspectratio]{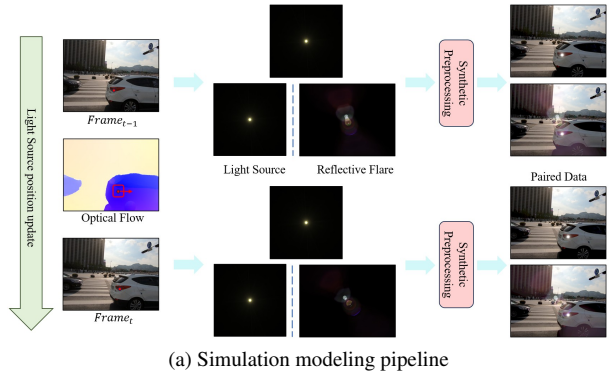} 
    \caption{Simulation modeling pipeline}\label{fig:overall_pipeline}
  \end{subfigure}\hfill
  \begin{subfigure}[t]{0.50\textwidth}
    \centering
    \includesvg[height=4.50cm,keepaspectratio]{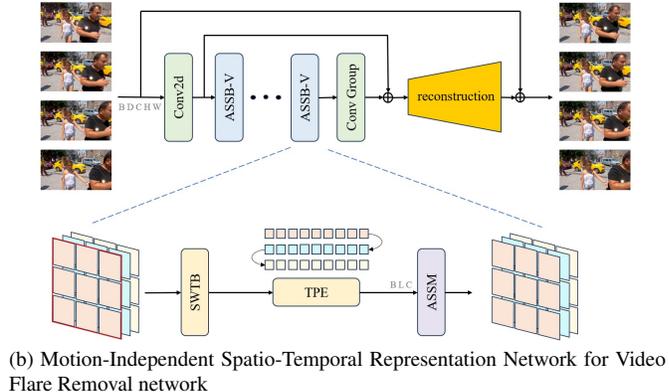} 
    \caption{Motion-Independent Spatio-Temporal Representation Network for Video Flare Removal network}\label{fig:overall_mivf}
  \end{subfigure}
  \caption{Dynamic Flare Synthesis Pipeline and video flare removal network.}\label{fig:overall_architecture}
\end{figure*}

\noindent\textbf{Reflective Flare.}
In dynamic flare variation, reflective flares exhibit more complex motion and appearance changes. 
Previous methods~\cite{dai2022flare7k, 2024Flare7K} typically rely on manual adjustments via software plugins to simulate limited variations of reflective flares. 
However, such methods are difficult to control and often produce monotonous flare dynamics. 
In our approach, we draw inspiration from Hullin~\cite{hu2011flare_rendering}, adopting a dynamic rendering strategy for reflective flares. 
By simulating multiple physically significant optical components, our method enables high-quality and photorealistic rendering of reflective flare patterns.

\noindent\textbf{Optical Flow–Based Dynamic Synthesis Strategy.}
As shown in Figure~\ref{fig:overall_pipeline}, both the static scattering and reflective flare synthesis require the light source positions as input. 
To capture realistic source motion in videos, we estimate the frame-wise positions of the light source using a strategy based on optical flow.

Given a video sequence
$\{I_0, I_1, \dots, I_t, \dots\}$,
we randomly sample an initial point on the first frame $I_0$ to represent the light source position 
$P_0^L(x, y)$. 
Based on the prior that the light source generally moves consistently with the motion direction of scene objects, 
we estimate the optical flow 
$O_{t-1 \to t}$ 
between adjacent frames 
$I_{t-1}$ and $I_t$ following FlowFormer++~\cite{shi2023flowformer++}, 
and compute the source position at time step $t$ as:

\begin{equation}
    P_t^L(x, y) = P_{t-1}^L(x, y) + O_{t-1 \to t}(x, y)
\end{equation}

\smallskip
For the scattering flare, its position is assumed to approximately coincide with the light source location, 
with a small random offset $t$ introduced experimentally:

\begin{equation}
    P_t^S(x, y) = P_t^L(x, y) + t, 
    \quad t \sim \mathcal{U}(0, 15)
\end{equation}

For the reflective flare, on the imaging plane, its position 
$P^R(x_1, y_1)$ 
always lies on the line connecting the light source 
$P^S(x_2, y_2)$ 
and the optical center 
$P^C(x_3, y_3)$,$\exists \lambda \in \mathbb{R}$,
which satisfies the linear constraint:

\begin{equation}
    \quad 
    \left( x_3 - x_1,\, y_3 - y_1 \right) 
    = \lambda \cdot 
    \left( x_2 - x_1,\, y_2 - y_1 \right)
\end{equation}

Through the above process, we compute the complete sequence of moving light source positions for a video scene. 

\subsection{Dataset}
We employed the dynamic flare synthesis pipeline to generate a large-scale video flare dataset. 
Clean (flare-free) video content was drawn from the REDS training set as the primary scene material. From the 240 REDS scenes, 
we sampled every 8th frame to form a new video scene, yielding 1,500 video scenes in total. To enhance data diversity, unique 
parameter configurations were used for reflective flare synthesis and rendering in each scene, and light source positions were 
sampled randomly. For synthesizing paired flare-degraded images, we principally followed Dai \etal~\cite{dai2022flare7k} that both the clean 
images and the flare templates were linearized via inverse gamma correction for linear composition, and the composited images 
were subsequently re-applied with gamma correction. The 1,500 scenes were split into 1,200 scenes for training and 300 scenes 
for testing. Because scattering (veiling) flare exhibits little variation across video frames, and to better enable the network 
to learn removal of more complex reflective flares, we additionally injected paired scattering flare into the test set as 
references for light source positions, and applied dynamic processing to these light source RGB values following Wu's method. 
Due to limited computational resources and to reduce the difficulty of video training, all frames were downsampled from the 
1280×720 pixels to 320×240 pixels.

\subsection{Video Flare Removal Network}
Following the recent progress in Mamba-based vision modeling~\cite{guo2025mambairv2, li2024videomamba}, 
we propose a hybrid spatio-temporal restoration network, named \textbf{MIVF}
(Motion-Independent Spatio-Temporal Representation Network for Video Flare Removal). 
Unlike conventional image restoration tasks, we primarily employ attention mechanisms for spatial feature modeling, 
while the Mamba-based \textit{Attention-State Space Module (ASSM)} is utilized for complex spatio-temporal dependencies. 
To better adapt the Mamba structure to video restoration scenarios, 
we design a variant named \textbf{ASSB-V}(Attention-State Space Block for Video), 
which enables efficient and accurate video flare removal.

\noindent{Overall Framework}.Given an input video 
sequence $I^{LQ} \in \mathbb{R}^{T\times H\times W\times C}$, 
where T, H, W and C are the video length, height, width and channel,the model 
outputs a restored sequence $I^{HQ} \in  \mathbb{R}^{T\times H\times W\times C}$ of 
the same resolution, thereby achieving the objective of video flare removal.
As illustrated in Figure~\ref{fig:overall_architecture}, the network comprises 
three main components: (1) a shallow feature extraction module; (2) a multistage 
spatio-temporal modeling backbone (stacked ASSB-V blocks); (3) a reconstruction head. 
Specifically, we first extract shallow features through a convolutional stem:  
$F^S \in \mathbb{R}^{T \times H \times W \times C}$.  
These features are then fed into multiple stacked Attention-State Space Blocks (ASSB-V) 
for sequential spatial and temporal modeling:
\begin{itemize}
    \item \textbf{Spatial Branch:} 
    Each frame is refined through 
    shifted-window transformer block (SWTB) 
    followed by a convolutional feed-forward network (ConvFFN), 
    producing detailed spatial features 
    $F^D \in \mathbb{R}^{T \times H \times W \times C}$.

    \item \textbf{Temporal Branch:} 
    A linear complexity temporal modeling operation is then performed 
    based on the Attention-State Space Module, 
    which captures long-range temporal dependencies and motion consistency, 
    yielding spatio-temporally coherent features 
    $F^T \in \mathbb{R}^{T \times H \times W \times C}$.
\end{itemize}
Finally, the decoding head reconstructs the video sequence through residual aggregation and convolutional layers, yielding a high quality video output. To facilitate comparison with other video restoration networks, our overall training framework adopts a conventional video restoration pipeline. The training is optimized using the Charbonnier loss, defined as:
\begin{equation}
    \mathcal{L} = \sqrt{\|\hat{I} - I^{HQ}\|^2 + \epsilon^2}
\end{equation}
where $\hat{I}$ is the restored sequence, $I^{HQ}$ is the ground-truth clean sequence, and $\epsilon$ is a small constant set to $10^{-3}$.

\noindent\textbf{Attentive State-Space Block}.
To extend the image restoration framework of Mamba to the video domain,
we redesign its Attentive State-Space Block (ASSB) into a video-adaptive structure, termed ASSB-V.
Unlike the original spatial-only design, ASSB-V enables efficient spatio-temporal dependency modeling under a unified state-space formulation while preserving spatial fidelity and linear time complexity.

Given the input video features$F^{S} \in \mathbb{R}^{T\times H\times W\times C}$, each frame is independently processed by a Shifted-Window Transformer Block with per-head relative positional bias. This design confines the quadratic attention cost within local spatial windows 
while maintaining cross window communication in a Swin style manner. Subsequently, a depthwise-separable convolutional feed-forward network (ConvFFN) is employed to inject convolutional inductive bias and enhance texture consistency:
\begin{equation}
F^{D}_{t} = \mathrm{ConvFFN}(\mathrm{SW\!-\!MHSA}(F^{S}_{t})),
\end{equation}
The resulting features $F^D \in \mathbb{R}^{T \times H \times W \times C}$ serve as refined spatial representations for subsequent temporal-state modeling.

\noindent\textbf{Temporal Aware State-Space Modeling.}
To jointly capture spatial and temporal correlations, 
we flatten the refined spatio-temporal features into a unified sequence:
\begin{equation}
F^{D}_{seq} \in \mathbb{R}^{B\times (T H W)\times C}.
\end{equation}
The sequence is processed by Attentive State-Space Module (ASSM), 
whose hidden dynamics follow an attentive variant of the linear state-space equation\cite{guo2025mambairv2}:
\begin{equation}
h_i = A h_{i-1} + B x_i, \quad
y_i = (C + P_i)h_i + D x_i,
\end{equation}
where $P_i$ represents the adaptive \textit{prompt} term derived from semantic routing, 
allowing non-causal global interaction with linear complexity.
To further embed temporal awareness into the state recurrence, 
we introduce an \textbf{Adaptive Temporal Positional Encoding (TPE)} prior to sequence modeling:
\begin{equation}
\tilde{F}^{D}_{t} = F^{D}_{t} + \alpha \cdot E^{temp}_{t},
\end{equation}
where $E^{temp}_{t}\!\in\!\mathbb{R}^{C}$ denotes a learnable temporal embedding
and $\alpha$ is a trainable scaling coefficient.
This adaptive temporal modulation enriches the latent dynamics 
with explicit frame-order sensitivity while maintaining causal consistency.
Meanwhile, the temporal encoding $E^{temp}_t$ ensures frame-order stability 
and prevents routing discontinuity during motion or illumination variations.
This design yields content yet temporally consistent token routing.

This unified design bridges attention and state‑space modeling within a single framework, 
enabling efficient long‑range temporal reasoning.
\section{Experiments}

\begin{figure*}[t]
  \centering
  \includesvg[width=0.95\linewidth]{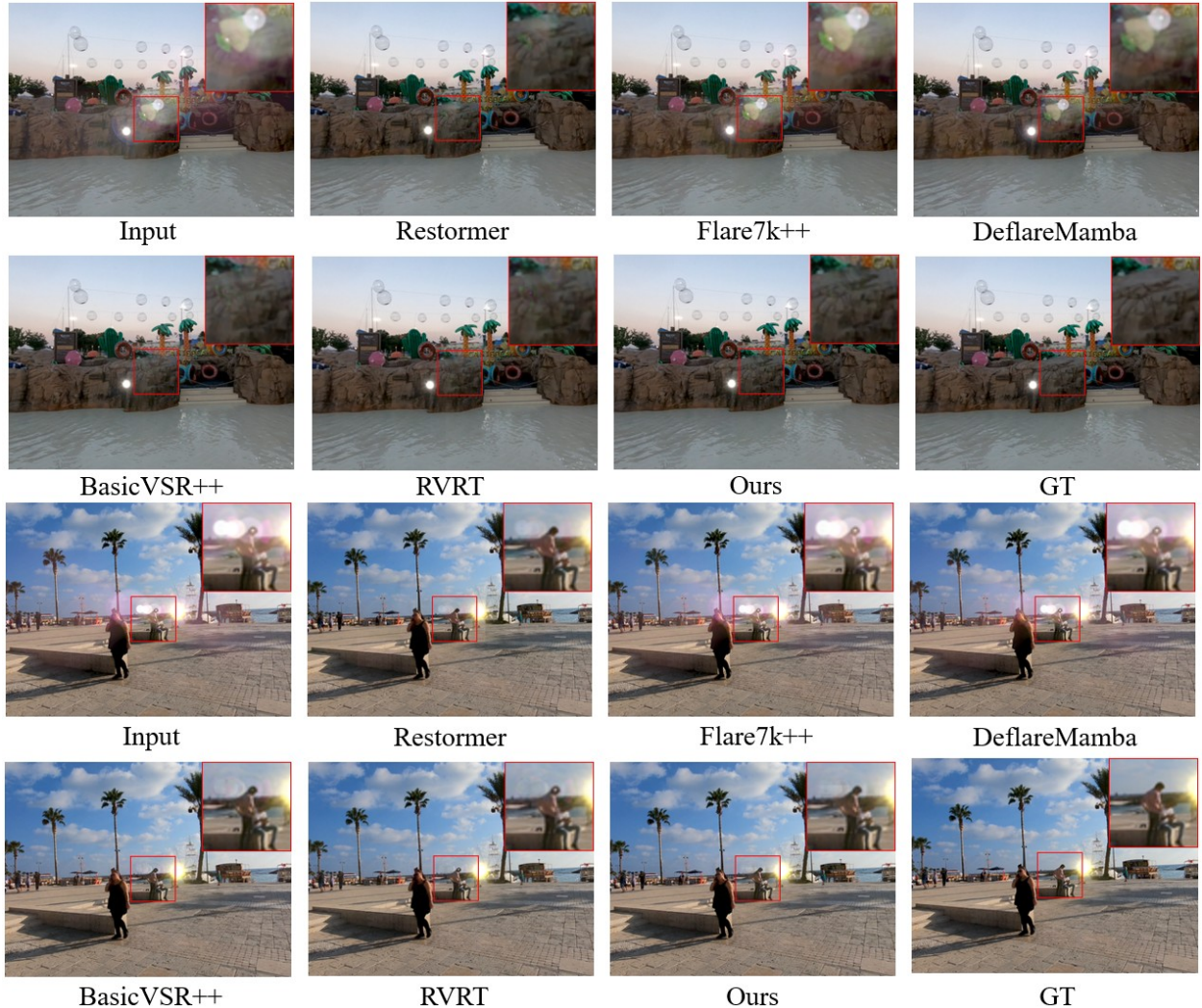}
  \caption{Visual comparison of different removal methods on synthetic flare video frames. In simulated data, our method removes glare more precisely without introducing additional artifacts.}\label{fig:syntheticDataPerformance}
\end{figure*}

\begin{table*}[t]
  \centering
  \caption{Comparison with state-of-the-art image and video restoration methods on our synthetic dataset.\dag\ indicates methods using official pre-trained weights. The best results are \textbf{bold}.}\label{tab:comparisonMethod}
  \footnotesize
  \begin{tabular}{lccccccc}
    \toprule
    Metric & \makecell{Input} & \makecell{BasicVSR++~\cite{chan2022basicvsr++}\\\textit{CVPR 2022}} & \makecell{RVRT~\cite{liang2022recurrent}\\\textit{NeurIPS 2022}} & \makecell{Restormer~\cite{zamir2022restormer}\\\textit{CVPR 2022}} & \makecell{Flare7K++\dag~\cite{2024Flare7K}\\\textit{TPAMI 2024}} & \makecell{DeflareMamba\dag~\cite{huang2025deflaremamba}\\\textit{MM 2025}} & \makecell{MIVF\\(\textbf{Ours})} \\
    \midrule
    PSNR$\uparrow$   & 28.82 & 40.59 & 40.22 & 41.10 & 29.65 & 30.58 & \textbf{41.54} \\
    PSNR-M$\uparrow$ & 31.32 & \textbf{36.71} & 36.51 & 36.00 & 31.74 & 32.15 & \textbf{36.71} \\
    SSIM$\uparrow$   & 0.9436 & 0.9836 & 0.9813 & 0.9849 & 0.9491 & 0.9533 & \textbf{0.9861} \\
    \bottomrule
  \end{tabular}
\end{table*}

\begin{figure*}[t]
  \centering
  \includesvg[width=0.95\linewidth]{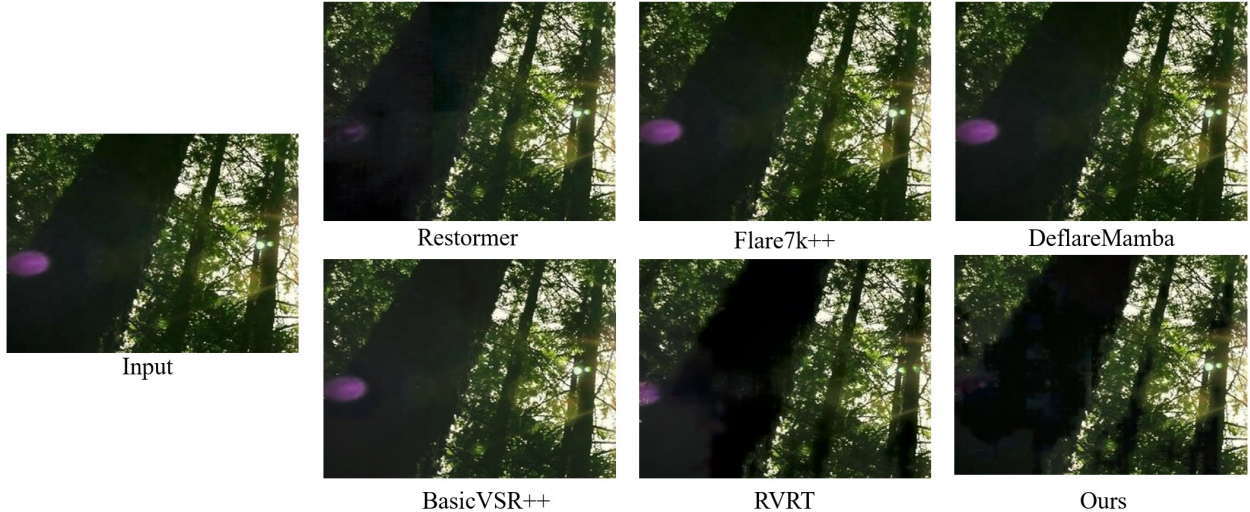}
  \caption{Visual comparison of different removal methods on real-world flare video frames. On real-world data, our approach achieves the best visual quality by effectively eliminating glare while preserving more scene details.}\label{fig:realDataPerformance}
\end{figure*}

\noindent\textbf{Experimental setting.} \
Our network is trained on the proposed video Flare dataset. 
Following common practices, we apply data augmentations such as random horizontal and vertical flips. 
Due to computational constraints, input frames are cropped into $128 \times 128$ patches during training. 
To further validate the effectiveness of our method, we also train and evaluate several existing video restoration architectures as well as image-based flare removal models. 
For certain flow-based networks that employ downsampling operations, 
the input resolution is adjusted to $160 \times 160$ to maintain architectural consistency.

To ensure a fair comparison, all models are trained using the AdamW optimizer, 
and the CosineAnnealingWarmRestarts scheduler is adopted for learning rate adjustment. 
Training is conducted on NVIDIA A800 GPUs. 
All models are comprehensively evaluated on the corresponding test sets, 
and the quantitative results are summarized in Table~\ref{tab:comparisonMethod}.

\noindent\textbf{Evaluation on Synthetic Data.} \
We construct a synthetic dataset containing 300 scenes to evaluate different restoration methods. 
This dataset serves as a controlled benchmark, allowing a quantitative comparison of model 
performance under various flare conditions.

\noindent\textbf{Evaluation on Real-World Data.} \
To assess the generalization ability of our synthesis-based approach, 
we collect a set of real-world videos from the Internet that contain prominent flare artifacts. 
These videos primarily feature moving scenes with strong reflective highlights, 
providing a suitable testbed for evaluating video flare removal.
Since obtaining ground truth counterparts for these real scenes is infeasible, 
the evaluation in this part focuses mainly on the visual quality achieved 
by different methods.

\subsection{Comparison with Previous Work}
To demonstrate the superiority of video-based approaches, we retrain several state-of-the-art 
video restoration networks under our experimental settings. To evaluate the effectiveness of 
each method, we first conduct experiments on the synthetic video dataset and report standard 
metrics such as PSNR and SSIM.
Following the design in Dai \etal\cite{2024Flare7K}, we additionally employ the PSNR-M metric 
to specifically assess the restoration performance in flare affected regions.
As illustrated in Figure~\ref{fig:syntheticDataPerformance}, our video-based model achieves the best overall 
results among all compared methods.
\begin{figure}[b]
  \centering
   \includesvg[width=0.95\linewidth]{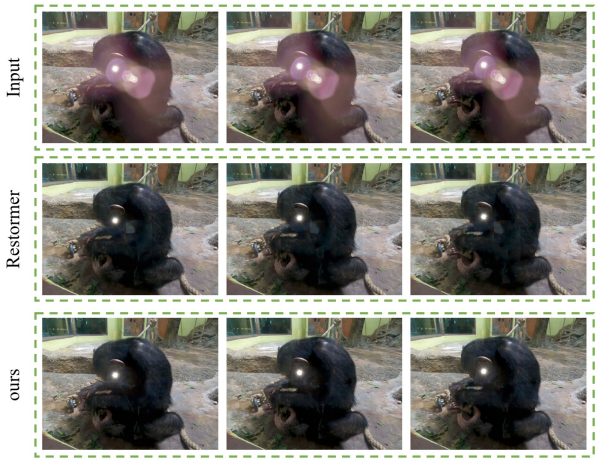}
   \caption{Visual comparison between image methods and video methods on real-world flare video frames.}\label{fig:differenceImageAndVideo}
\end{figure}

As shown in Table~\ref{tab:comparisonMethod}, under flare conditions, video restoration 
methods that rely on temporal alignment perform worse than image-based approaches. 
This observation is consistent with the previously discussed motion independence between the scene 
and flare. In contrast, our method leverages the physically perceptual motion-independent 
trait of 
flare, avoiding multi-frame alignment and learning directly from video sequences. Compared with 
the second-best method, our approach improves the PSNR by 0.36 dB. 
The SSIM score is 0.0012 higher than the second-best method.

\noindent\textbf{Visual Comparison between image method and video method.} 
Since the degradation process of flare signals is temporally continuous, image-based methods can suppress 
flare to some extent but tend to produce flickering and artifacts.
As shown in Figure~\ref{fig:differenceImageAndVideo}, our video-based approach achieves superior visual performance compared with the 
current state-of-the-art image restoration methods.
By incorporating the Mamba module for long sequence temporal information aggregation, our method ensures 
spatio-temporal consistency during restoration.

\subsection{Ablation Study}
\noindent\textbf{Effectiveness of the Synthetic Generation Method.}
We also evaluated several previous image-based methods on real-world datasets.
Most of these image-based approaches primarily model scattering flare, and therefore perform 
poorly in many reflection dominant flare scenarios.
In contrast, our method explicitly considers the physical properties of flare, particularly its 
motion independence from the surrounding scene, enabling it to better mimic the motion characteristics 
and dynamic variations of flare in real-world conditions.
As shown in Figure~\ref{fig:realDataPerformance}, models retrained on our newly collected dataset achieve 
significantly better restoration results than those obtained by previous methods.

\begin{table}[h]
  \centering
  \caption{Ablation study on the Spatio-Temporal Modeling. We evaluate the impact of the Spatio-Temporal Modeling on flare removal performance. The best results are \textbf{bold}.}\label{tab:ablation}
  \small
  \begin{tabular}{@{}lccc@{}}
    \toprule
    \multicolumn{1}{c}{Model Variant} & \multicolumn{1}{c}{PSNR$\uparrow$} & \multicolumn{1}{c}{PSNR-M$\uparrow$} & \multicolumn{1}{c}{SSIM$\uparrow$} \\
    \midrule
    w        & \textbf{41.54} & \textbf{36.71} & \textbf{0.9861} \\
    w/o         & 41.01 & 36.22 & 0.9853 \\
    \bottomrule
  \end{tabular}
\end{table}

\noindent\textbf{ASSM Module with Spatio-Temporal Modeling.}
To evaluate the effectiveness of our temporal modeling, we removed the temporal encoding and the ASSM module 
from the original ASSB architecture.
The quantitative results are presented in Table~\ref{tab:ablation}.
Experimental results demonstrate that the proposed components lead to a significant improvement in 
PSNR and other metrics, indicating that ASSB-V effectively refines scene features by performing 
long-sequence temporal modeling.

\section{Conclusion}

In this work, we investigate the problem of video flare and analyze its general patterns of variation in real-world scenarios. 
Based on the \textbf{motion-independence} between flare and scene, we design a physically-informed dynamic flare synthesis pipeline and introduce a new video flare dataset, Vflare.
To address the motion inconsistency between flare and scene content, we propose the Motion-Independent Spatio-Temporal Representation Network for Video Flare Removal \textbf{(MIVF)}, which leverages long sequence modeling to mitigate interference between flare motion and scene motion across frames.
Experiments demonstrate the effectiveness of both our flare synthesis and removal methods.

\noindent\textbf{Discussion.}
Although we have analyzed various motion behaviors and appearance changes of flare, certain conditions remain unexplored.
When the light source is located outside the scene, the corresponding flare motion pattern has not been thoroughly studied.
Factors such as camera rotation or defocusing can also affect the spatial trajectory and appearance of flare, but these subtle influences are not yet modeled in our current synthesis pipeline.
Moreover, our method does not always perform well on real-world data which particularly in low light scenes, where the flare removal quality still lags behind.
We attribute this to the inherent bias of the synthetic dataset.
Future work may focus on improving reflection flare modeling and exploring domain adaptation techniques to bridge the gap between synthetic and real-world scenarios.


{
    \small
    \bibliographystyle{ieeenat_fullname}
    \bibliography{main}
}


\end{document}



\clearpage
\setcounter{page}{1}
\maketitlesupplementary

\begin{strip}
\centering
  {\large Junqiao Wang \quad Yuanfei Huang \quad Hua Huang}\\[0.0pt]
  \large School of Artificial Intelligence, Beijing Normal University\\[0.0pt]
  {\tt\small 202321081032@mail.bnu.edu.cn \quad \{yfhuang, huahuang\}@bnu.edu.cn}
  \vspace{12pt}
\end{strip}

\setcounter{section}{0}
\setcounter{figure}{0}
\setcounter{table}{0}
\setcounter{equation}{0}

\section{Details on Reflective Flare Principle}\label{sec:reAnalysis}
The variation of reflective flare constitutes the most critical component of the overall dynamic behavior of flare. The specific motion relationship between light and flare can be described by the optical transfer matrix model\cite{lee2013practical}. Under a small-angle approximation (the first-order Maclaurin expansion), the interaction between light rays and the optical system can be represented as matrix multiplication.

In the 2D meridional plane, a light ray $l$ can be expressed as a two dimensional vector $\mathbf{r} = [r ~\theta]^{\mathrm{T}}$. The interaction between the ray and each optical interface is characterized by a transfer matrix $\mathbf{M}_i$. For different types of transmission, the corresponding matrices are given as follows:

\begin{equation}
\mathbf{T}_i =
\begin{pmatrix}
1 & d_i \\
0 & 1
\end{pmatrix}
\quad
\end{equation}

\begin{equation}
\mathbf{R}_i =
\begin{pmatrix}
1 & 0 \\
\dfrac{n_1 - n_2}{n_2 R_i} & \dfrac{n_1}{n_2}
\end{pmatrix}
\quad
\end{equation}

\begin{equation}
\mathbf{L}_i =
\begin{pmatrix}
1 & 0 \\
\dfrac{2}{R_i} & 1
\end{pmatrix}
\quad
\end{equation}

$d_i$ is a positive displacement to the next interface at interface $i$, $n_1$ and $n_2$ the refractive indices at interface i and $R_i$ its lens radius. We denote $\mathbf{T}_i$ the translation matrix with displacement $d_i$, $\mathbf{R}_i$ the refraction, and $\mathbf{L}_i$ the reflection matrices at interface $i$.

\begin{figure}[t]
  \centering
   \includesvg[width=0.95\linewidth]{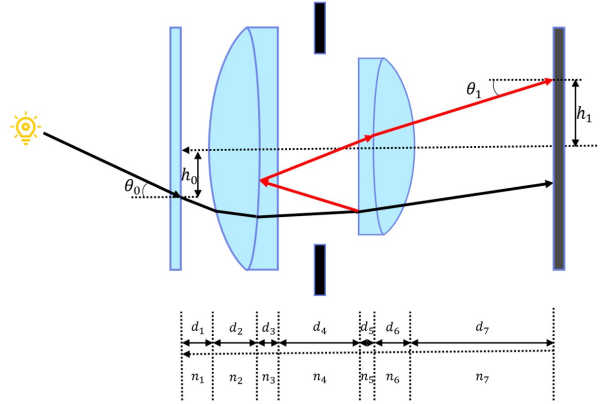}
   \caption{Flare matrix formulation. $d$ and $n$ indicate optical distances between optical interfaces and refractive indices.}\label{fig:re}
\end{figure}

As illustrated in Figure~\ref{fig:re}, The complex refraction and reflection behavior of a single ray can thus be described as follows, and $\mathbf{D}_{n}=\mathbf{T}_{n} \mathbf{R}_{n}$:

\begin{equation}
\mathbf{H} = \mathbf{D}_{7} \mathbf{D}_{6} \mathbf{D}_{5} \mathbf{D}_{4} \mathbf{T}_{3} \mathbf{L}_{3}^{-1} \mathbf{T}_{3} \mathbf{R}_{4}^{-1} \mathbf{T}_{4} \mathbf{L}_{4}\mathbf{D}_{3} \mathbf{D}_{2} \mathbf{T}_{1}
\end{equation}

Therefore, the relationship between the reflective flare and the incident light can be expressed as:

\begin{equation}
\begin{pmatrix}
h_1 \\
\theta_1
\end{pmatrix}
=
\mathbf{H}\
\begin{pmatrix}
h_0 \\
\theta_0
\end{pmatrix}
\end{equation}

As the incident angle of light changes, the reflective flare also varies correspondingly.
Moreover, due to the presence of the aperture structure inside the camera, the reflective flare is likely to undergo additional phenomena such as clipping, flipping, or partial obstruction.

\section{More Details on Dynamic Flare Synthesis}\label{sec:pipline}

\subsection{Light Source Position}
In the synthesis process, the initial position of the light source is randomly assigned.
However, to ensure that the light source remains within the field of view and does not move out of frame with the motion of the scene, we define a constrained region for its initialization.
Specifically, the light source is randomly placed within the central two-thirds of the image area.
It should be noted that the optical flow–based estimation is not perfectly accurate; in a small number of cases, errors may occur due to object occlusions.
We have manually filtered out part of the data that exhibited such severe inconsistencies.

\begin{figure}[t]
  \centering
   \includesvg[width=1.00\linewidth]{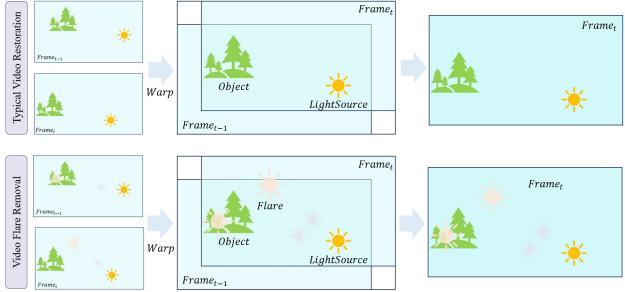}
   \caption{The difference between Video Flare Removal and Typical Video Restoration.}\label{fig:diff1}
\end{figure}

\subsection{Reflective Flare Generation}
Reflective flare represents the most prominent component in the dynamic variations of flare. 
The shape and position of the reflective flare are primarily determined by the location of the light source and the configuration of the camera’s optical system.
Given a known light source position, the rendering of reflective flare in our method follows the approach described in~\cite{hu2011flare_rendering}.

Specifically, we employ a fixed camera lens parameter module based on the \textbf{Nikon (28–75\,mm)} lens. 
The lens model consists of 29 optical interfaces, each characterized by the following parameters:
\begin{itemize}
    \item $r$, radius of curvature (mm).
    \item $d$, distance to the next interface (mm).
    \item $n$, refractive index.
    \item $f$, flatness indicator.
    \item $w$, thickness scaling factor for rendering.
    \item $h$, semi-aperture (mm).
    \item $c$, coating thickness (nm).
\end{itemize}

To generate more diverse flare effects, we randomly vary additional parameters such as the aperture diameter, light source intensity, and coating quality.
In future work, we plan to explore a broader range of flare types by varying lens parameters and other optical settings.

\subsection{Flare Synthesis}
During the synthesis process, to enable the model to better learn the motion behavior of reflective flare, we did not include the real flare dataset \textbf{Flare-R} from Dai~\etal\cite{2024Flare7K}. when synthesizing scattered flare.
The Flare-R dataset contains a large number of samples that include both scattered and reflective flare.
Since these two types of flare cannot be effectively separated. Combining them would disrupt the physical motion characteristics of the overall flare pattern.

When simulating scattered flare, we followed the method of Wu~\etal\cite{wu2021train} but did not include strong streak-shaped scattering effects, as excessively intense scattered flare could severely interfere with the synthesis of reflective flare.
Specifically, we simulated scattered flare using simple scratches and dirt patterns, and generated various common RGB light source and flare images by modifying the corresponding RGB transformation matrix.

In constructing the dataset, to encourage the model to focus more on learning the dynamic variation of reflective flare while still introducing interference from scattered flare, the final synthetic pairs consisted of two types of images:
(1) composite images of scattered flare and background.
(2) composite images of scattered flare, reflective flare, and background.

\section{Discussions on Alignment-based Mothods}\label{sec:AlignModel}
We conducted a detailed analysis of the failure cases observed in certain video-based models for this problem.
More specifically, as illustrated in Figure~\ref{fig:diff1}, typical low-level vision tasks such as deblurring and denoising benefit from temporal alignment, since features shared across consecutive frames can enhance scene reconstruction.
However, in the case of video flare removal, the independence between flare patterns and scene motion leads to inconsistencies after temporal alignment.
The aligned frames may contain flare artifacts corresponding to different timestamps, which consequently weakens the restoration capability.

\begin{figure}[b]
  \centering
   \includesvg[width=1.00\linewidth]{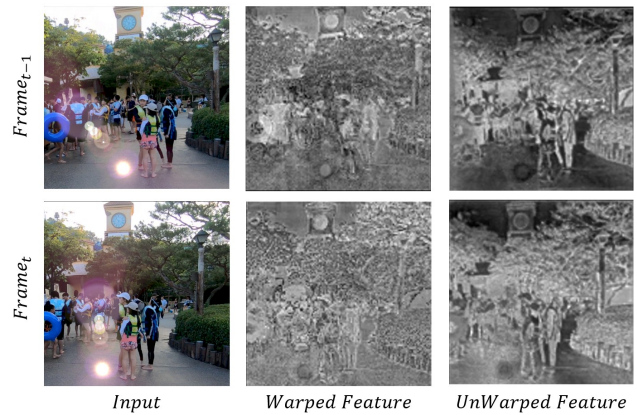}
   \caption{We compared the feature maps obtained using alignment-based and non-alignment methods.
The non-aligned feature extraction results were obtained by removing the RFR component from the RVRT~\cite{liang2022recurrent} framework, while ensuring that the extracted feature maps were of the same network depth.
The results indicate that, due to the independence between scene motion and flare motion, the algorithm fails to accurately estimate optical flow, leading to inferior restoration performance of alignment-based methods in this type of task.}\label{fig:featuremap}
\end{figure}

As shown in Figure~\ref{fig:featuremap}, the feature maps clearly illustrate this phenomenon that flare artifacts vary across different alignment stages, indicating that the temporal alignment becomes less effective under such conditions.

\begin{figure*}[t]
  \centering
  \includesvg[width=1.00\linewidth]{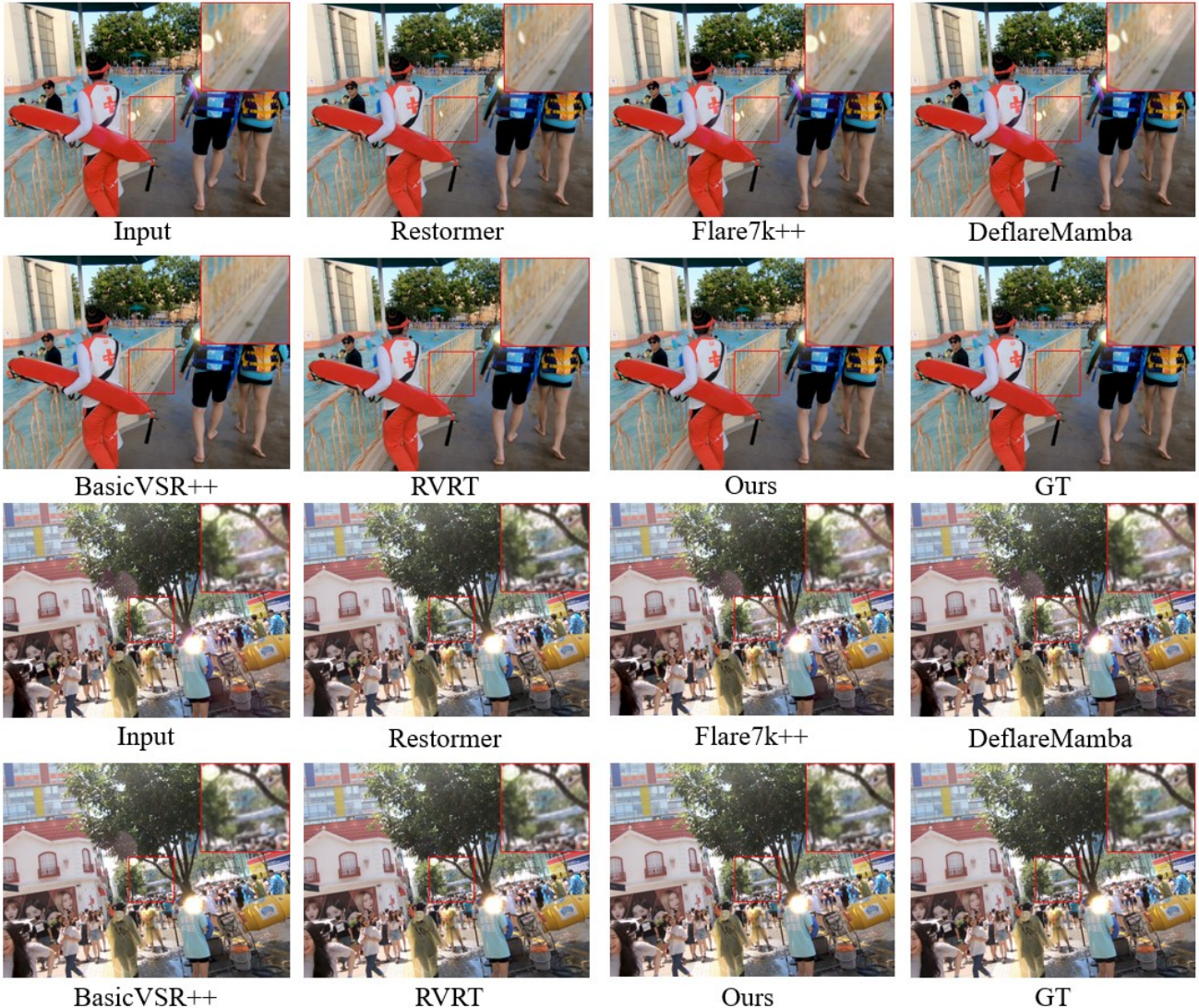}
  \caption{Visual comparison of different removal methods on synthetic flare video frames. On synthetic data, our approach achieves the best visual quality by effectively eliminating flare while preserving more scene details.}\label{fig:synDataPerformance}
\end{figure*}

\section{More Details on  Experimental Setup}\label{sec:setup}
Due to computational resource limitations, in addition to downsampling the synthetic dataset for training, we further adjusted the input size and several training parameters.
During training, we uniformly used sequences of 4 frames as input.
During inference, since our model extracts features through long-sequence information aggregation, it requires 4-frame inputs for segment-wise inference, whereas other models take the entire sequence as input.
The block size used during inference was kept consistent with that used in training.

\begin{figure*}[h]
  \centering
  \includesvg[width=1.00\linewidth]{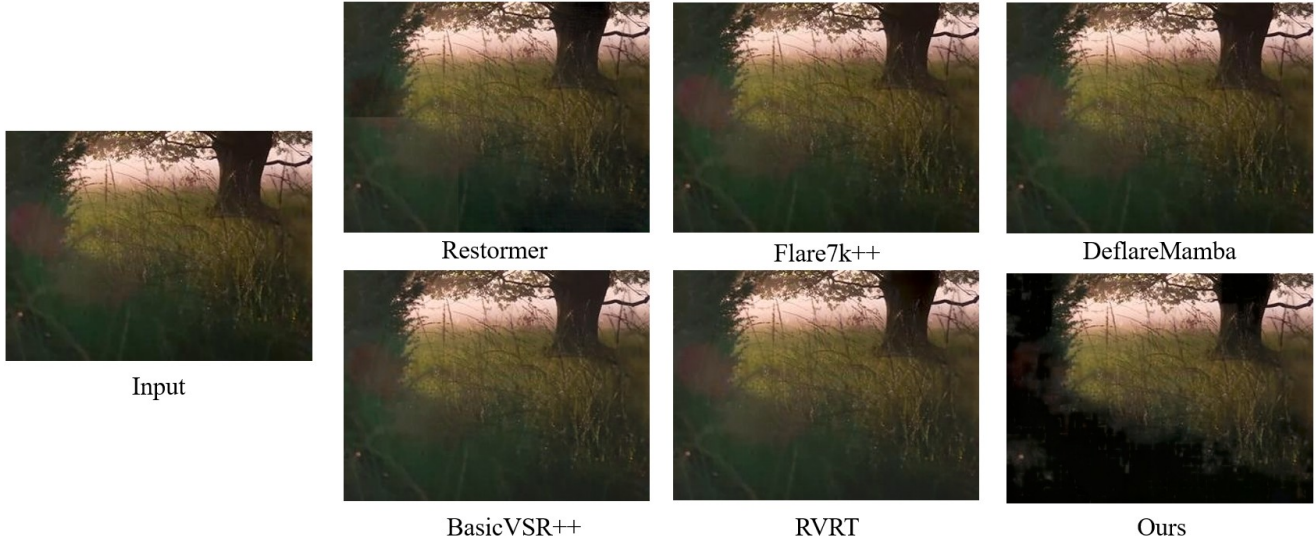}
  \caption{Visual comparison of different removal methods on real-world flare video frames. On real-world data, our approach also performs well on various halo-like artifacts}\label{fig:realDataPerformance}
\end{figure*}

\begin{table}[h]
  \centering
  \caption{Video comparison on the synthetic flare dataset using temporal metrics. Higher tPSNR and lower tLPIPS~\cite{krizhevsky2012imagenet} indicate better spatio-temporal consistency.}
  \label{tab:supp_vertical_comparison}
  \begin{tabular}{@{}lcc@{}}
    	\toprule
    	\textbf{Method} & \textbf{tPSNR$\uparrow$} & \textbf{tLPIPS$\downarrow$} \\
    \midrule
    BasicVSR++~\cite{chan2022basicvsr++} & 39.67 & 0.0067 \\
    RVRT~\cite{liang2022recurrent} & 39.13 & 0.0072 \\
    \textbf{MIVF (Ours)} & \textbf{40.54} & \textbf{0.0059} \\
    \bottomrule
  \end{tabular}
\end{table}

\section{Video Temporal Consistency Evaluation}\label{sec:spatio-temporal}
In the main text, we presented a visual comparison between video-based flare removal methods and image-based flare removal methods.
Here, we specifically adopt two metrics tPSNR and tLPIPS to evaluate the spatio-temporal consistency of video restoration methods, as shown in Table~\ref{tab:supp_vertical_comparison}.

\section{More Results on Synthesis and Real-World Dataset}\label{sec:realData}
We also evaluated the effectiveness of our dataset and flare removal method in additional scenes.
In addition to the reflective flare spots shown in the main text, our approach also performs well on various halo-like artifacts.
Compared with the previous Flare7K++ image dataset, our method achieves superior performance in reflective flare removal.
Furthermore, when compared with other state-of-the-art restoration methods, our approach consistently outperforms them in both flare removal and detail preservation.

{
    \small
    \bibliographystyle{ieeenat_fullname}
    \bibliography{main}
}
